\documentclass[journal]{IEEEtran}

\usepackage{orcidlink}
\usepackage{amsmath,amssymb,amsfonts}
\usepackage{algorithmic}
\usepackage{graphicx}
\usepackage{textcomp}
\usepackage{float}
\usepackage{caption}
\usepackage{multirow}
\usepackage{mathtools}
\usepackage{bm}
\usepackage{adjustbox}
\usepackage{subfigure}
\usepackage[table]{xcolor}
\usepackage{soul}
\usepackage{gensymb}
\usepackage{authblk}
\usepackage[utf8]{inputenc}
\usepackage{booktabs}
\usepackage{nameref}
\usepackage{placeins}
\usepackage[ruled,vlined]{algorithm2e}
\SetKw{Return}{return}
\usepackage{hyperref}
\usepackage{siunitx} % Recommended for units
\usepackage[T1]{fontenc}

\ifCLASSINFOpdf
\else
\fi

\begin{document}

\title{Model-Based and Neural-Aided Approaches for Dog Dead Reckoning}

\author{Gal~Versano~\orcidlink{0009-0007-7766-2511},
        Itai Savin~\orcidlink{0009-0007-6911-5091},
        Itzik~Klein~\orcidlink{0000-0001-7846-0654}
        
\thanks{I. Klein, G. Versano and I. Savin are with the Hatter Department of Marine Technologies, University of Haifa, Israel.}}% <-this % stops a space
\maketitle

\begin{abstract}
Modern canine applications span medical and service roles, while robotic legged dogs serve as autonomous platforms for high-risk industrial inspection, disaster response, and search and rescue operations. For both, accurate positioning remains a significant challenge due to the cumulative drift inherent in inertial sensing. To bridge this gap, we propose three algorithms for accurate positioning using only inertial sensors, collectively referred to as dog dead reckoning (DDR). To evaluate our approaches, we designed DogMotion, a wearable unit for canine data recording. Using DogMotion, we recorded a dataset of 13 minutes. Additionally, we utilized a robotic legged dog dataset with a duration of 116 minutes.  Across the two distinct datasets we demonstrate that our neural-aided methods consistently outperform  model-based approaches, achieving an absolute distance error of less than  10\%. Consequently, we provide a lightweight and low-cost positioning solution for both biological and legged robotic dogs. To support reproducibility, our codebase and associated datasets have been made publicly available.
\end{abstract}

\begin{IEEEkeywords}
Dead Reckoning, Dogs, Robotic Legged Dogs, Inertial Sensors, Deep Learning  
\end{IEEEkeywords}

\section{Introduction}
\noindent Recent advancements in deep learning (DL) have opened new frontiers for analyzing animal behavior and movement \cite{boteju2020deep, mao2023deep, hussain2023deep, kasnesis2022deep}.  Focusing on canine applications, DL methods were applied to identify orthopedic and neurological disorders by analyzing the canine gait~\cite{palez2025canine} and for activity recognition (like sitting, walking, or standing)~\cite{nolasco2025dog, cetintav2025decoding}.\\
\noindent
Autonomous navigation for legged (quadrupedal) robots has recently evolved from simple path planning to more dynamic approaches. Modern research, led by groups such as \cite{hutter2024wild} and \cite{wisth2023vilens}, focuses on combining sensors like LiDAR and depth cameras. By using vision-based reinforcement learning and multi-sensor state estimation, these robots can navigate unstructured environments such as collapsed buildings or dense forests, where traditional 2D mapping methods do not work well \cite{li2026muse}. Recently, a work employing visual language model to reduce the localization drift in challenges environments was proposed~\cite{cheng2024navila}.  Wisth et al.~\cite{wellhausen2023artplanner} presented ArtPlanner, a sampling-based navigation planner that uses learned foothold evaluation and motion costs to enable reliable real-time navigation of quadrupedal robots in complex environments. In a recent work, Elnoor er al. \cite{elnoor2024pronav} proposes ProNav, a navigation method that uses  sensor data  to estimate terrain traversability of robot dog. Multiple inertial measurement units and encoders were fused in an extended Kalman filter  to achieve accurate positioning with a cost of additional hardware~\cite{yang2023multi}.
\noindent
While robots can be programmed with specific, predictable motion patterns, such constraints cannot be applied to biological subjects. Unlike engineered robotic platforms with rigid links and known kinematic constraints, animal motion is highly non-rigid, intermittent, and context-dependent. This unpredictability motions makes classical model-based estimation challenging, traditional algorithms often fail to account for the complex biomechanical variations across different breeds, sizes, and environmental terrains, making them sensitive to sensor placement and gait variability. \\
\noindent 
In the related field of human movement, namely pedestrian dead reckoning (PDR)~\cite{wang2022recent,klein2025pedestrian}, has  leveraged  inertial sensors for the positioning task. First model-based approaches used accelerometers and biomechanical models to estimate step length with high precision \cite{ho2016step, shin2007adaptive}. To maintain accurate orientation, these systems typically integrate gyroscopes and magnetometers through attitude and heading reference systems such as the Madgwick filters \cite{madgwick2010efficient}. Foot-mounted systems further utilize Zero-Velocity Update (ZUPT) techniques to periodically reset accumulated integration errors whenever the sensor is stationary during the stance phase \cite{wagner2022reevaluation, li2023adaptive}. \\
\noindent 
Deep learning has recently emerged as a powerful tool for overcoming the limitations of model-based PDR.  Neural networks were employed to estimate the step length and velocity directly from high-dimensional sensor data \cite{Klein2020StepNet}. Furthermore, end-to-end architectures, such as have demonstrated the ability to reconstruct entire trajectories by learning the underlying physics of motion directly from raw inertial measurements \cite{li2023learning,asraf2021pdrnet, zhang2023pedestrian, liu2020tlio, chen2018ionet}. \\
\noindent
In robotic applications, periodic trajectories emulating pedestrian motion are applied to mimic PDR approaches. Those include quadrotors~\cite{hurwitz2024deep,shurin2020qdr} and wheeled mobile platforms~\cite{ etzion2023morpi,sahoo2025morpi}. \\
\noindent 
Despite the success of PDR-like methods in human and robotic applications, a significant gap remains in robust inertial frameworks suitable for both biological dogs and legged dog robots. To bridge this gap, we propose three inertial-based algorithms for accurate positioning, collectively referred to as dog dead reckoning (DDR). Leveraging the potential of DL methods to handle complex movement patterns and nonlinear dependencies, two of the three proposed algorithms are based on neural networks. 
Despite the main differences between biological dogs and robotic legged dogs (the former exhibit a wider variety of gait parameters), our DDR methods are designed to cope both types. \\
\noindent
Our main contributions are summarized as follows:
\begin{enumerate}
    \item \textbf{Dog Dead Reckoning (DDR)}: A set of algorithms for accurate positioning of both biological dogs and robotic legged dogs based only on low-cost inertial sensors. 
    \item \textbf{Model-Based DDR}: Development of  a model-based  DDR pipeline that adapts the Weinberg step-length estimation approach and the Madgwick algorithm for walking direction estimation. Combining the two enables trajectory estimation.
    \item \textbf{Deep Learning-Aided DDR}: Derivation of two different deep learning-aided DDR approaches. The first estimates the velocity magnitude and direction using residual networks. The second employs an encoder transformer architecture to estimate the heading angle. In both cases, the predicted velocity is integrated over time to reconstruct the trajectory. 
    \item \textbf{DDR Repository}: To ensure reproducibility, provide a standardized benchmark for the community, and stimulate further research, we have made our codebase and a novel dataset publicly available via a \href{https://github.com/ansfl/DDR}{GitHub repository}. 
\end{enumerate}
To evaluate our proposed approaches we  designed DogMotion, a wearable unit for canine data recording. Using DogMotion we recorded a dataset of 13 minutes. In addition, we employed a robotic legged dataset with a duration of 116 minutes.  Across the two distinct datasets we demonstrate that our deep learning methods consistently outperform the model-based approach with an absolute distance error of less than 10\%.
\noindent The rest of the paper is organized as follows: 
Section~\ref{sc:PF} presents the inertial navigation and PDR equations of motion. Section~\ref{propose_app} describes our proposed approach and Section~\ref{sec:andr} presents the results. Finally, Section~\ref{sec:conc} concludes the paper.
%
%חיישנים אינרצאיליים
%\noindent Inertial Measurement Unit (IMU) are ideal for localization because they are lightweight, low-power, and do not require external infrastructure \cite{xu2022novel}. However, the double integration of noisy acceleration and angular velocity leads to rapid drift, where small biases accumulate into massive position errors over time. To combat this, data-driven models are increasingly used to learn motion priors that constrain the search space of the navigation solution \cite{shurin2020qdr, cohen2024inertial}. These networks can be trained to predict short-term displacement or uncertainty-related parameters, which are then fused with model-based filters, such as the Extended Kalman Filter (EKF), to provide a more stable and robust estimation \cite{chen2018ionet, qiu2023airimu}.
%
\section{Problem Formulation}\label{sc:PF}
\subsection{Inertial Navigation Systems}\label{sc:fw}
The INS equations of motion derive the navigation state (position, velocity, and orientation) by processing inertial sensor measurements relative to a known set of initial conditions. \\
Let the accelerometer measurements (specific force vector), $\boldsymbol{f}_{ib}^{b}$, expressing in the body frame be denoted by
\begin{equation}\label{eq:fimu}
\boldsymbol{f}_{ib}^{b}=
\left [
\textit{f}_{\textit{x}} \quad \textit{f}_{\textit{y}} \quad \textit{f}_{\textit{z}} \right ]^{T},
\end{equation}
and  the gyroscopes measurements (angular velocity vector), $\boldsymbol{\omega}_{ib}^{b}$, expressed in the body frame be denoted by 
\begin{equation}\label{eq:wimu}
 \boldsymbol{\omega}_{ib}^{b} = \left [\omega_x \quad \omega_y \quad \omega_z \right ]^{T} 
\end{equation}
When using low-cost gyroscopes that cannot detect the Earth's rotation, or during short operational durations, the Earth's turn rate can be neglected. In these scenarios, the INS equations of motion are simplified to \cite{Titterton2004}:
\begin{eqnarray}
 \dot{\textbf{p}}^{n} &=& \textbf{v}^{n} \label{eq:INSEOM1}\\
 \dot{\textbf{v}}^{n} &=& \textbf{T}^{n}_{b}\textbf{f}_{ib}^{b} + \textbf{g}^{n}\label{eq:INSEOM2}\\
 \dot{\textbf{T}}^{n}_{b} &=& \textbf{T}^{n}_{b}\boldsymbol{\Omega}_{ib}^{b}\label{eq:INSEOM3}
 \end{eqnarray}
 where $\textbf{p}^{n}$ is the position vector expressed in the local navigation frame, $\textbf{v}^{n}$ is the velocity vector expressed in the navigation frame, $\textbf{g}^{n}$ is the local gravity vector expressed in the navigation frame,  $\boldsymbol{\Omega}_{ib}^{b}$ is a skew-symmetric form of the angular velocity vector and  ${\dot{\textbf{T}}}^{n}_{b}$ is the transformation matrix from body to navigation frame given by:
\begin{equation}\label{eq:DCM}
\textbf{T}^{n}_{b}=\begin{bmatrix}c_{\theta}c_{\psi}&s_{\phi}s_{\theta}c_{\psi}-c_{\phi}s_{\psi}&c_{\phi}s_{\theta}c_{\psi}+s_{\phi}s_{\psi}\\c_{\theta}s_{\psi}&s_{\phi}s_{\theta}s_{\psi}+c_{\theta}c_{\psi}&c_{\phi}s_{\theta}s_{\psi}-s_{\phi}c_{\psi}\\-s_{\theta}&s_{\phi}c_{\theta}&c_{\phi}c_{\theta}\end{bmatrix}
\end{equation}
where $s_{x}$ is the sine of $x$ and $c_{x}$ is the cosine of $x$.
\subsection{Pedestrian Dead Reckoning} \label{PDRform}
The model-based PDR algorithm cycle is comprised of four primary modules: step detection, step-length estimation, heading determination, and position updating.
After each cycle, the 2D pedestrian position is updated by~\cite{klein2025pedestrian}:
\begin{eqnarray}\label{eq:pdrpos}
	\textit{x}_{k} & = & \textit{x}_{k-1}+\textit{s}_{k}\cos\psi_{k}\\
	\textit{y}_{k} & = & \textit{y}_{k-1}+\textit{s}_{k}\sin\psi_{k}
\end{eqnarray}
where $\psi_{k}$ is the user’s walking direction angle, $textit{s}_{k}$ is the estimated step size, $k$ is the current epoch, and $\textit{x}_{k}$ and $\textit{y}_{k}$
are the two-dimensional position components.\\
One of the common approaches to estimate the step-length is the  Weinberg's approach \cite{weinberg2002using}. Based on the specific force magnetite, from \eqref{eq:fimu},  the Weinberg's formula is \cite{Klein2020StepNet}:
\begin{equation}\label{eq:steplength2}
        s_w = k_w \left (f_{mag,max} - f_{mag,min} \right )^{1/4}
\end{equation}
where $k_w$ is the Weinberg gain, $s_w$ is the Weinberg-based step-length, $f_{mag,max}$ is the maximum value of the specific force during the step interval, and $f_{mag,min}$ is the minimum value of the specific force during the step interval.
\section{Proposed Approach} \label{propose_app}
\noindent
We offer two families of algorithms for the positioning of dogs and robotic legged dogs using only inertial sensors. Motivated by the extensive literature on PDR, we propose model-based and deep learning-aided approaches to estimate the dog's position robustly and accurately. In the latter, we introduce two methods for estimating the dog walking direction: one based on residual networks and the other on a transformer encoder.
\subsection{Model-Based DDR}
\begin{figure*}[h]
    \centering
    \includegraphics[width=1.0\linewidth]{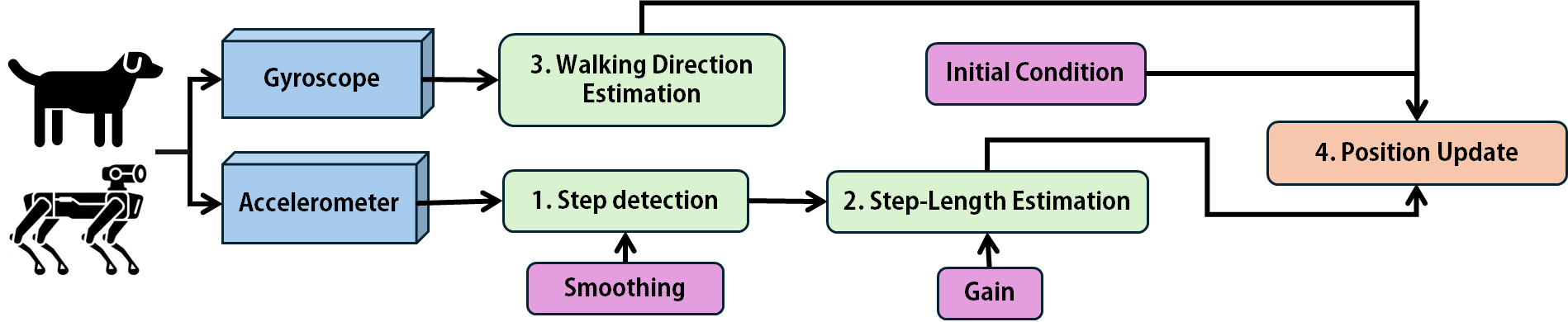}
    \caption{Flow diagram of our proposed model-based DDR approach.}
    \label{fig:schematic_pde}
\end{figure*}

\noindent
We follow the PDR guidelines and employ a four stage model-based DDR approach, as illustrated in Figure~\ref{fig:schematic_pde} and described below. \\
\noindent
\textbf{1. Step Detection}.  The dogs steps are identified from accelerometer measurements by applying a peak detection algorithm to the specific force magnitude instances~\cite{harle2013survey}. Prior to applying the algorithm, the data was smoothed using a moving average filter.  \\
\noindent
The magnitude of the specific force vector \eqref{eq:fimu} at time $k$ is
\begin{equation} \label{eq:fmag}
    {f}_{mag,k} = \sqrt{f_{x,k}^2+f_{y,k}^2+f_{z,k}^2}
\end{equation}
The mean of the specific force magnitude throughout the trajectory is defined by
\begin{equation} \label{eq:fmagmean}
    \bar{f}_{mag} = \frac{1}{n} \sum^{n}_{k=1}{f}_{mag,k}
\end{equation}
where $n$ is the number of samples. Next, The mean of the specific force magnitude \eqref{eq:fmagmean} is subtracted from the magnitude of the specific force vector \eqref{eq:fmag}
\begin{equation} \label{eq:fmagmean1}
    {f}_{m,k} = {f}_{mag,k} - \bar{f}_{mag} 
\end{equation}
Finally, the standard deviation (STD) of the specific force magnitude, after reducing its mean, is calculated by
\begin{equation} \label{eq:fmagstd}
    \sigma_{f}= \left [
    \frac{1}{n} \sum^{n}_{k=1}({f}_{m,k} - \bar{f}_{mag})^2
\right ]^{\frac{1}{2}}    
\end{equation}
Finally, two parameters are required to implement this peak detection approach: 1) the minimum time interval between steps, defined empirically; and 2) the minimum peak height, which represents the threshold for a valid peak. We employ $1.5\sigma_{f}$ as this minimum threshold. \\
\noindent
\textbf{2. Step-Length Estimation}.  The step-length between two successive peaks is calculated by applying the Weinberg approach \eqref{eq:steplength2} on the dog's detected peaks:
\begin{equation}\label{eq:Wdog}
        s_d = k_d \left (f_{mag,max} - f_{mag,min} \right )^{1/4}
\end{equation}
where $k_d$ is the Weinberg gain, $s_d$ is the Weinberg-based step-length. We note that the Weinberg algorithm is a biomechanical approach, based on the inverted pendulum model, developed for humans we apply it here for dogs. While the physical principles of locomotion are similar, there are fundamental differences due to humans being bipeds (two-legged) and dogs being quadrupeds (four-legged). Nevertheless, we assume that these differences are compensated when determining the Weinberg gain value. \\
\textbf{3.Walking Direction Estimation}.
We adopt the Madgwick algorithm for attitude and heading estimation \cite{madgwick2010efficient}. The Madgwick filter is quaternion-based orientation estimator. The algorithm propagates the orientation using the gyroscope measurements and applies a correction term driven by accelerometer and magnetometers updates. As magnetometers are highly sensitive to the environment, we employ only accelerometer corrections. Specifically, the quaternion kinematics are propagated by:
\begin{equation}
{\mathbf{\dot q}_{w,t}} = \frac{1}{2}\,\mathbf{q_{w,t-1}}\otimes {\boldsymbol{\omega}}^b_{ib,t}
\label{eq:madgwick_qdot}
\end{equation}
where \(\mathbf{q_{w}}=[q_0,q_1,q_2,q_3]^\mathsf{T}\) is the unit quaternion representing the rotation from the body frame to the navigation frame, \(\boldsymbol{\omega}^b_{ib}\) is the angular velocity vector measured by the gyroscope in the
body frame, and \(\otimes\) denotes the quaternion multiplication. \\
\noindent To reduce gyroscope bias, the Madgwick filter adds a nonlinear correction term. This term is computed using gradient descent and estimated attitude to align the predicted angular velocity vector with the measured one from the accelerometer. The corrected quaternion dynamics are:
\begin{equation}
{\dot {\mathbf{q}}_{est,t}} = {\mathbf{\dot q}_{w,t}} 
-\beta\,\frac{\nabla f\!}{\lVert \nabla f \!\rVert}
\label{eq:madgwick_corrected}
\end{equation}
where \(\beta\) is a gain that controls the strength of the correction and $\nabla f$ is the gradient of the objective function $f(\mathbf{q}_t)$ with respect to the quaternion components, indicating how $f$ changes for a small change in $\mathbf{q}_t$. \\
\noindent The orientation at the current time step is obtained by numerically integrating the corrected quaternion rate $\dot{\mathbf{q}}_{est, t}$ \eqref{eq:madgwick_corrected}. Using first-order discrete-time integration with a sampling period $\Delta t$, the orientation update is expressed as:
\begin{equation}
\mathbf{q}_{t} = \hat{\mathbf{q}}_{t-1} + \dot{\mathbf{q}}_{est, t} \Delta t 
\label{eq:madgwick_integration}
\end{equation}
where $\hat{\mathbf{q}}_{t-1}$ is the estimated orientation from the previous time step. \\
\noindent By using the estimated unit quaternion $\mathbf{\hat q_t}=[q_1,q_2,q_3,q_4]^\mathsf{T}$, the heading angle, indicating the walking direction, is:
\begin{equation}
\psi_t = \mathrm{atan2}\left [ 2(q_3 q_4 - q_1 q_2),\; 2(q_1^2 + q_4^2)-1\right ]
\label{eq:yaw_from_quaternion}
\end{equation}
\textbf{4.Position Update}.
Given an initial 2D position of the dog $(x_{k-1}, y_{k-1})$, the dog's position is updated recursively:
\begin{subequations}
    \begin{align}
        x_{k} &= x_{k-1} + s_d \cos (\psi_k) \\
        y_{k} &= y_{k-1} + s_d \sin (\psi_k)
    \end{align}
\end{subequations}
where $s_d$ is calculated in \eqref{eq:Wdog} and the walking direction $\psi_k$ is estimated in \eqref{eq:yaw_from_quaternion}. 
To summarize, our proposed model-based DDR is presented in Algorithm~\ref{alg:ddr}. 
\begin{algorithm}[t]
\caption{Model-Based DDR Algorithm.}
\label{alg:ddr}
\DontPrintSemicolon
\SetKwInOut{Input}{Input}
\SetKwInOut{Output}{Output}
\Input{
Inertial measurements $\{\mathbf{a}^b_t,\boldsymbol{\omega}^b_t\}_{t=1}^{T}$ at rate $f_s$\\
Sampling period $\Delta t = 1/f_s$\;
Weinberg gain $G$\;
Peak detection parameters: threshold $\gamma$, minimum step period $T_{\min}$\;
Madgwick gain $\beta$\;
Initial position $\mathbf{p}_0=[x_0,y_0]^\mathsf{T}$\;
Initial quaternion $\hat{\mathbf{q}}_0=[q_{1,0},q_{2,0},q_{3,0},q_{4,0}]^\mathsf{T}$\;
}
\Output{
Estimated 2D trajectory $\{\mathbf{p}_k\}_{k=0}^{K}$\;
}
\BlankLine
\textbf{Initialize:} $\mathbf{p}_0=[x_0,y_0]^\mathsf{T}$, $\hat{\mathbf{q}}_0$, $k\leftarrow 0$\;
\BlankLine
\For{$t=1$ \KwTo $T$}{
     $\mathbf{q}_{\omega,t}\leftarrow [0,\omega^b_{x,t},\omega^b_{y,t},\omega^b_{z,t}]^\mathsf{T}$\;
    $\dot{\mathbf{q}}_{\omega,t}\leftarrow \frac{1}{2}\,{\hat {\mathbf{q}}_{t-1}} \otimes \mathbf{q}_{\omega,t}$\;
    Compute $\nabla f(\hat{\mathbf{q}}_{t-1},\mathbf{a}^b_t)$ from gravity alignment error\;
    $\dot{\mathbf{q}}_{\text{est},t}\leftarrow \dot{\mathbf{q}}_{\omega,t} - \beta \frac{\nabla f}{\|\nabla f\|}$\;
     $\hat{\mathbf{q}}_{t}\leftarrow {\hat{\mathbf{q}}_{t-1}} + \dot{\mathbf{q}}_{\text{est},t}\Delta t$\;
     $\hat{\mathbf{q}}_{t}\leftarrow \hat{\mathbf{q}}_{t}/\|\hat{\mathbf{q}}_{t}\|$\;
     $\psi_t \leftarrow \mathrm{atan2}\!\Big(2(q_{3,t}q_{4,t}-q_{1,t}q_{2,t}),\, 2(q_{1,t}^2+q_{4,t}^2)-1)\Big)$\;
     $f_t \leftarrow \|\mathbf{a}^b_t\|$\;
     $f_{\max}\leftarrow \max(f_{\max}, f_t)$\;
     $f_{\min}\leftarrow \min(f_{\min}, f_t)$\;
     \If{isPeak($f_t,\gamma$) \textbf{and} $(t\Delta t - t_{\text{last}})\ge T_{\min}$}{
         $t_{\text{last}} \leftarrow t\Delta t$\;
         $k \leftarrow k+1$\;
         $\Delta f \leftarrow \max(f_{\max}-f_{\min},\,0)$\;
         $s_k \leftarrow G \cdot \sqrt[4]{\Delta f}$\;
         $\psi_k \leftarrow \psi_t$\;
         $x_k \leftarrow x_{k-1} + s_k\cos(\psi_k)$\;
         $y_k \leftarrow y_{k-1} + s_k\sin(\psi_k)$\;
         Store $\mathbf{p}_k=[x_k,y_k]^\mathsf{T}$\;
     }
}
\Return $\{\mathbf{p}^n_k\}_{k=0}^{K}$\; 
\end{algorithm}
\subsection{Deep Learning-Aided DDR with ResNets}
\begin{figure*}[h]
    \centering
    \includegraphics[width=0.8\linewidth]{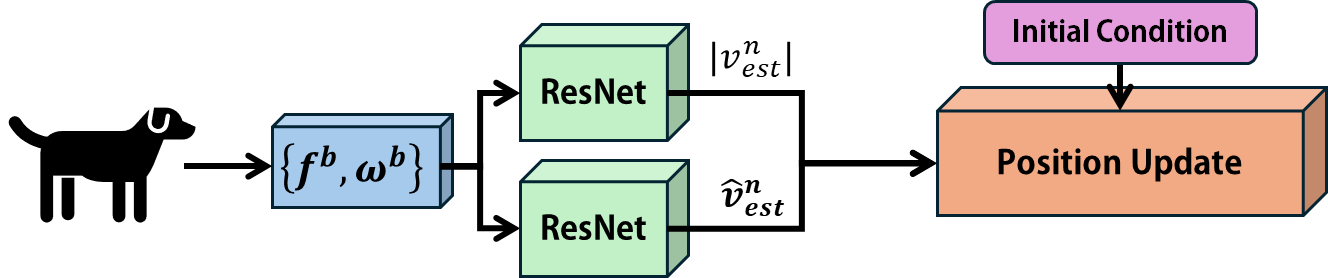}
    \caption{Our proposed deep learning-aided DDR using a ResNet backbone.}
    \label{fig:DogNet}
\end{figure*}

Model-based DDR approaches face several significant limitations.  A major drawback of model-based DDR is the requirement for pre-calibrated empirical gains to estimate step-length, which must often be determined for each specific dog prior to operation. Furthermore, these gains are extremely sensitive to user dynamics (such as walking speed).  The position solution can be updated only between successive peaks, which limits the update rate.
To overcome these limitations, we offer a model-free deep learning approach to estimate the dog's position through the estimation of its velocity and walking direction as shown in Figure~\ref{fig:DogNet}.
To this end, we split the positioning task into two specialized deep learning models:
\begin{enumerate}
    \item  \textbf{Velocity Estimation}: This network focuses on how fast the dog is moving. It looks at the patterns in the accelerometer data to identify the dog's pace and translates those vibrations into a forward velocity.
    \item \textbf{Walking Direction Estimation}: This network focuses on where the dog is heading. It analyzes the rotation data to track the dog's heading, even during sharp turns and outputs the walking direction.
\end{enumerate}
Once the magnitude velocity and the direction are regressed,  the 2D position of dog is updated by:
\begin{equation}
    \mathbf{p^n_{k+1}}=\mathbf{p^n_k}+\mathbf{\hat v^n_{est,k}} \cdot |v^n_{est,k}| \cdot \Delta t
\end{equation}
where $|v^n_{est,k}|$, $\mathbf{\hat v^n_{est}}$ are the estimation of the magnitude velocity and direction in time $k$, respectively, $\mathbf{p^n_k}$ is the position vector in time $k$, and $\Delta t$ is the update rate.
\subsubsection{Backbone Architecture}
\noindent
The network architecture is based on a 1D adaptation of the ResNet-18 \cite{he2016deep} architecture. 
Since our input is a stacked sequence of 1D inertial measurements, we replace the original 2D convolutional with a 1D convolution that operates along the temporal axis and accepts a $6$-channel window, corresponding to tri-axial accelerometer and gyroscope readings.
To capture the complex temporal dynamics of movement, the input is processed through a series of residual blocks containing 1D convolutional layers, where skip connections are employed to facilitate gradient flow. These features are subsequently passed through fully connected layers with ReLU activations to introduce nonlinearities, ultimately outputting a regressed value.\\
\noindent
The resulting backbone preserves the ResNet-18 stage configuration $(2,2,2,2)$ %with  \togal{check:} BasicBlocks
, while operating entirely in 1D. Following the residual stages, we flatten the result and used a fully connected regression head to predict the velocity/direction. This design follows common practice in neural inertial navigation for mapping raw IMU windows to motion estimates \cite{herath2020ronin, asraf2021pdrnet}.\\
\noindent
This backbone architecture, with all parameters, is used for both regression tasks. The only difference is in the output layer. For the velocity estimation, the output is a $2 \times 1$ vector representing the directional components of the velocity while for the walking direction the output is a scalar of the direction.  
\subsubsection{Loss Functions and Training}
\noindent 
To optimize the network for velocity estimation, the mean squared error (MSE) loss function is employed during the training process:
\begin{equation}
L(y_i, \hat{y}_i) = \frac{1}{n} \sum_{i=1}^{n} (y_i - \hat{y}_i)^2
\end{equation}
where $y_i$ represents the ground truth (GT) velocity magnitude, $\hat{y}_i$ is the predicted value, and $n$ denotes the number of items in the batch.\\
\noindent
For  walking direction regression network, we employ a cosine distance loss defined as:
\begin{equation}
    L(\theta)=1-cos(\theta)
\end{equation}
where
\begin{equation}
    cos(\theta)=\mathbf{\hat v^n_{est}} \cdot \mathbf{\hat v^n_{GT}}
\end{equation}
\noindent
The network parameters, including weights and biases denoted by the vector $\alpha$, are optimized using the Adaptive Moment Estimation (Adam) algorithm to ensure robust convergence. The update rule is defined as:
\begin{equation}
\alpha = \alpha - \eta \nabla_{\alpha} J(\alpha), \quad \alpha = [\omega \quad b]^T 
\label{update_rule}
\end{equation}
where $J(\alpha)$ is the loss function, which emphasizes that the parameters are optimized as a collective vector $\alpha$, $\eta$ denotes the learning rate, and $\nabla_{\alpha}$ is the gradient operator. \\
\noindent The initial learning rate is set to 0.009, with a learning rate scheduler configured to apply a reduction factor of 0.5 upon detecting a plateau in validation performance.The training process is executed over 200 epochs with a batch size of 2048.
Each input instance consists of a $6 \times w$ matrix, representing one second of synchronized data from the three-axis accelerometer and three-axis gyroscope sampled at $w$ Hz. This windowing approach allows the backbone architecture to learn the spatial-temporal correlations necessary for estimating the velocity magnitude and velocity direction and also to mitigate overfitting. The train parameters is sum up in Table \ref{hyper_table}.
\begin{table}[!h]
\centering
\caption{Hyperparameters for the ResNet backbone architecture.}
\resizebox{0.45\textwidth}{!}{ 
\begin{tabular}{|c|c|c|c|c|}
\hline
\textbf{Learning rate} & \textbf{Batch Size} & \textbf{Epoch} &\textbf{Window Size}  \\ \hline
0.009 & 2048 & 200 & 125 \\ \hline
\end{tabular}}
\label{hyper_table}
\end{table}
\subsubsection{Summary}

\noindent The entire deep-learning-aided DDR algorithm process shown in Algorithm~\ref{alg:dognet_position_update}.
\begin{algorithm}[h]
\caption{Deep learning-aided DDR using a ResNet backbone.}
\label{alg:dognet_position_update}
\DontPrintSemicolon
\SetKwInOut{Input}{Input}
\SetKwInOut{Output}{Output}

\Input{
IMU windows $\{\mathbf{X}_k\}_{k=0}^{K-1}$\\
Trained networks: magnitude velocity model $\mathcal{F}_{\text{vel}}$ and direction velocity model $\mathcal{F}_{\text{dir}}$\;
Initial position $\mathbf{p}^n_0 \in \mathbb{R}^2$ (e.g., $\mathbf{p}^n_0=[0,0]^\mathsf{T}$)\;
Time step $\Delta t$\;
}
\Output{
Estimated trajectory $\{\mathbf{p}^n_k\}_{k=0}^{K}$
}

\BlankLine
Initialize $\mathbf{p}^n_0$;\;

\For{$k \leftarrow 0$ \KwTo $K-1$}{
    Predict velocity magnitude:\;
    \quad $ |v^n_{\text{est},k}| \leftarrow \mathcal{F}_{\text{vel}}(\mathbf{X}_k)$ \;
    Predict velocity direction:\;
    \quad $\mathbf {\hat v^n_{est,k}} \leftarrow \mathcal{F}_{\text{dir}}(\mathbf{X}_k)$ \;

    Construct velocity vector:\;
    \quad $\mathbf{v}^n_{\text{est},k} \leftarrow \mathbf {\hat v^n_{est,k}} |v^n_{\text{est},k}|$\;
    
    Position update:\;
    \quad $\mathbf{p}^n_{k+1} \leftarrow \mathbf{p}^n_k + \mathbf{v}^n_{\text{est},k}\,\Delta t$\;
}

\Return $\{\mathbf{p}^n_k\}_{k=0}^{K}$\;
\end{algorithm}
\subsection{Deep Learning-Aided DDR with Transformer Encoders}
\begin{figure*}
    \centering
    \includegraphics[width=1.0\linewidth]{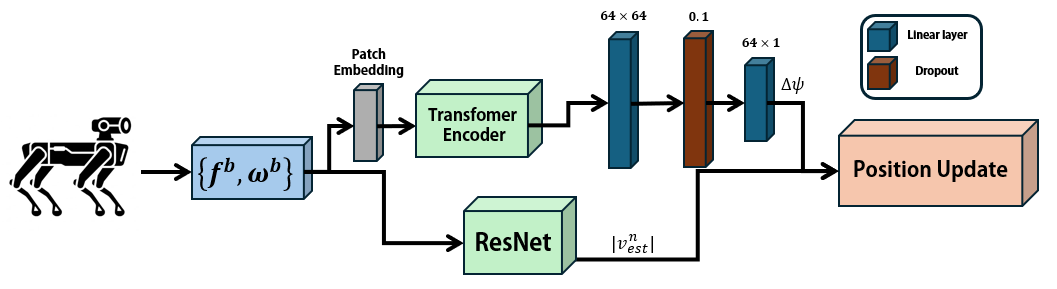}
    \caption{Our proposed deep learning-aided DDR using a hybrid ResNet and transformer encoder backbone.}
    \label{LRE_archi}
\end{figure*}

\noindent
We offer a different approach for the walking direction estimation. It includes three main building blocks as follows:
\begin{enumerate}
 \item \textbf{Embedding Layer}: The embedding layer employs a 1D convolutional layer with a kernel size of three to project the raw inertial sensor input from six channels to a latent dimension of $d_{model} = 64$. This convolutional approach allows the model to capture local temporal features within the time series—such as specific gait impacts or high-frequency vibrations—thereby enhancing the model’s expressive capacity compared to standard linear projections in points. The resulting features are scaled by $\sqrt{d_{model}}$ to ensure numerical stability during the subsequent attention operations.
 \item \textbf{Encoder Layer}: The core of the model consists of three stacked transformer encoder layers designed to extract long-range dependencies within the inertial data. Each layer contains a multi-head self-attention (MHSA) mechanism with four heads, which allows the model to simultaneously attend to different patterns in the motion sequence. The attention operation is formally defined as:
    \begin{equation}
        \text{Attention}(Q, K, V) = \text{softmax} \left( \frac{QK^T}{\sqrt{d_k}} \right) V
    \end{equation}
    where $Q, K$, and $V$ represent the query, key, and value matrices, respectively, and $d_k$ denotes the dimension of the key vectors. Following the attention block, a position-wise feed-forward network (FFN) with a hidden dimension of 128 is applied. Each sub-layer is wrapped in a residual connection followed by layer normalization to facilitate stable gradient flow. 
 \item \textbf{Regression Head:} The final representation is extracted from the latent state of the last time step in the sequence, providing a temporally aware summary of the entire motion window. This vector is passed through a fully connected regression head consisting of a hidden layer with 64 units and ReLU activation.
\end{enumerate}
Thus, the proposed network architecture utilizes a transformer-based framework to estimate the heading from inertial sensor data. The model receives a time window of one second, consisting of $w$ samples of synchronized data from a three-axis accelerometer and a three-axis gyroscope. To capture the complex temporal dynamics of movement, the input is processed through an encoder based transformer. The  output is passed through two fully connected layers with ReLU and Dropout function between them. A flowchart of the entire process is shown in Figure~\ref{LRE_archi}.
\subsubsection{Loss Function and Training}
\noindent 
To optimize the network for heading estimation, the MSE loss function is employed during the training process:
\begin{equation}
L(\psi_i, \hat{\psi}_i) = \frac{1}{n} \sum_{i=1}^{n} (\psi_i - \hat{\psi}_i)^2
\end{equation}
where $\psi_i$ is the GT heading displacement and $\hat{\psi}_i$ is the predicted heading displacement. \\
\noindent
The network parameters are optimized using the Adam algorithm to ensure robust convergence. The update rule is defined as shown in \eqref{update_rule}. The learning rate is set to 0.0025. The training process is executed over 200 epochs with a batch size of 1024. Each input instance consists of a $6 \times w$ matrix, representing one second of synchronized data from the three-axis accelerometer and three-axis gyroscope sampled at $w$ Hz. The train parameters are summarized in Table \ref{hyper_table_LRE}.
\begin{table}[!h]
\centering
\caption{Hyperparameters for the ResNet and transformer encoder backbone architectures.}
\resizebox{0.45\textwidth}{!}{ 
\begin{tabular}{|c|c|c|c|c|}
\hline
\textbf{Learning rate} & \textbf{Batch Size} & \textbf{Epoch} &\textbf{Window Size}  \\ \hline
0.0025 & 1024 & 200 & 100 \\ \hline
\end{tabular}}
\label{hyper_table_LRE}
\end{table}
\subsubsection{Position update}
\noindent 
Once we calculate the magnitude velocity and the heading estimation we can calculate the position of dog by:
\begin{equation}
    \mathbf{p^n_{k+1}}=\mathbf{p^n_k}+|v^n_{est,k}|\cdot [cos(\psi_{k-1}+\Delta\psi_{k}), sin(\psi_{k-1}+\Delta\psi_{k})] \cdot \Delta t
\end{equation}
where $|v^n_{est,k}|$, $\Delta\psi_{k+1}$ are the estimation of the magnitude velocity and heading displacement in time $k$, respectively. \\
\noindent 
The deep-learning aiding DDR approach is summarized in Algorithm~\ref{alg:LRE_position_update}.
\begin{algorithm}[h]
\caption{Deep learning-aided DDR using a hybrid ResNet and transformer encoder backbone.}
\label{alg:LRE_position_update}
\DontPrintSemicolon
\SetKwInOut{Input}{Input}
\SetKwInOut{Output}{Output}

\Input{
IMU windows $\{\mathbf{X}_k\}_{k=0}^{K-1}$\;
Trained models: $\mathcal{F}_{\text{mag}}$ and $\mathcal{F}_{\text{dir}}$ \;
Initial state: Position $\mathbf{p}^n_0 \in \mathbb{R}^2$ and Heading $\psi_0$\;
Time step $\Delta t$\;
}
\Output{
Estimated trajectory $\{\mathbf{p}^n_k\}_{k=0}^{K}$
}

\BlankLine
Initialize $\mathbf{p}^n_0$ and $\psi_0$\;

\For{$k \leftarrow 0$ \KwTo $K-1$}{
    $|v^n_{\text{est},k}| \leftarrow \mathcal{F}_{\text{mag}}(\mathbf{X}_k)$ \;
    $\Delta\psi_k \leftarrow \mathcal{F}_{\text{dir}}(\mathbf{X}_k)$ \;
    
    $\psi_k \leftarrow \psi_{k-1} + \Delta\psi_k$ \;
    
    $\mathbf{d}_k \leftarrow |v^n_{\text{est},k}| \cdot [\cos(\psi_k), \sin(\psi_k)]^\mathsf{T}$ \;
    $\mathbf{p}^n_{k+1} \leftarrow \mathbf{p}^n_k + \mathbf{d}_k \cdot \Delta t$ \;
}

\Return $\{\mathbf{p}^n_k\}_{k=0}^{K}$\;
\end{algorithm}

\section{Analysis and Results}\label{sec:andr}
\subsection{Datasets}
\noindent
To evaluate our proposed approach, we employ two datasets. Our dataset was recorded using two dog subjects, while the publicly available second dataset was collected using a legged robot~\cite{frey2026grandtour}.
\subsection{Our Dog Dataset} 
\begin{figure*}[h]
    \centering
    \includegraphics[width=1.0\linewidth]{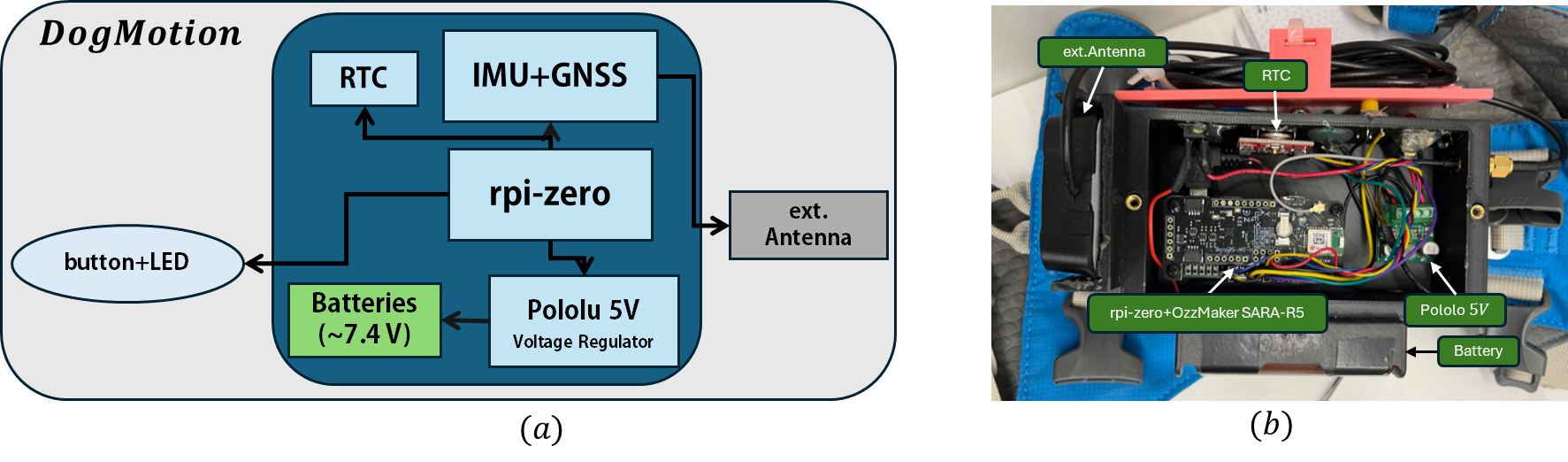}
    \caption{Our DogMotion device (a) Hardware connection and (b) Design and components.}
    \label{dog_mot_platform}
\end{figure*}

\begin{figure*}[h]
    \centering
    \includegraphics[width=1\linewidth]{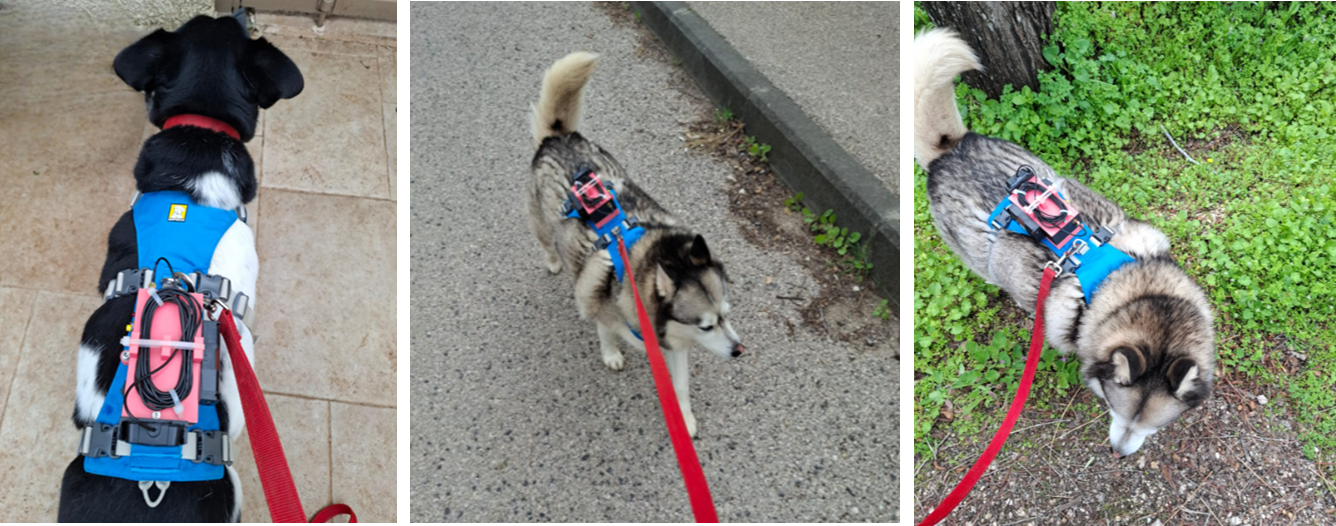}
    \caption{The dogs equipped with our DogMotion device  during field experiments.}
    \label{fig:dog_imu}
\end{figure*}

\noindent The data collection was performed using our own designed DogMotion device. It is a custom-integrated sensing unit based on the Raspberry Pi Zero microcomputer \cite{RPiZero}. The core of the system is the OzzMaker SARA-R5 LTE-M GPS \cite{OzzMakerSARA} shield, which provides inertial readings alongside GNSS psoition measurements.
To ensure temporal consistency a critical requirement for sensor fusion a real time clock (RTC) module was integrated into the architecture. This module maintains strict synchronization between the high-frequency inertial samples and the GNSS reference signals, preventing time-drift during extended recording sessions. The system is powered by a 7.4 [V] battery, providing a stable voltage for long-duration field trials. User interaction is managed via a button equipped with an LED indicator, which serves as the control interface to initiate and terminate data logging sessions and provides real-time status feedback. The DogMotion device is presented in Figure~\ref{dog_mot_platform}.
The experimental dataset was recorded using two dog subjects equipped with the DogMotion sensor suite as presented in Figure~\ref{fig:dog_imu}. 
The dataset comprises a total of 13 minutes of synchronized motion recordings across 12 different trajectories, as described in Table \ref{time_traj},
\begin{table}[h]
\centering
\caption{Recorded trajectories length and time in our dog dataset.}
\label{time_traj}

\begin{tabular}{|c|c|c|}
\hline
\textbf{Trajectory} & \textbf{Length [m]} & \textbf{Time {[}sec{]}} \\ \hline
1                   & 76.1              & 46                     \\ \hline
2                   & 41.1               & 33                     \\ \hline
3                   & 71.9                & 53                     \\ \hline
4                   & 95.6               & 65                     \\ \hline
5                   & 61.8               & 66                     \\ \hline
6                   & 79.8               & 50                     \\ \hline
7                   & 97.0               & 71                     \\ \hline
8                   & 118.3               & 87                     \\ \hline
9                   & 119.5              & 98                     \\ \hline
10                  & 100.8              & 76                     \\ \hline
11                  & 98.2               & 57                     \\ \hline
12                  & 103.9             & 62                     \\ \hline
\textbf{Average} &   \textbf{88.7} & \textbf{63.7} \\ \hline
\end{tabular}
\end{table}
\noindent 
The inertial readings was configured to sample at 125 [Hz], providing dense temporal data. In contrast, the GNSS GT was recorded at a frequency of 1 [Hz]. To facilitate supervised learning, the GNSS coordinates were interpolated and synchronized with the inertial time-steps, allowing for precise labeling of the velocity and direction vectors, the full specification of the inertial sensors is provided in Table \ref{spec_imu}.
\begin{table}[h]
\centering
\caption{Inertial Sensor Bias and Noise Density Specifications.}
\resizebox{\columnwidth}{!}
{
\label{spec_imu}
\renewcommand{\arraystretch}{1.3}
\begin{tabular}{|cc|cc|}
\hline
\multicolumn{2}{|c|}{\textbf{Gyroscope}} & \multicolumn{2}{c|}{\textbf{Accelerometer}} \\ \hline
\textbf{Bias} [\unit{\degree/h}] & \textbf{Noise} [\unit{\degree/s/\sqrt{Hz}}] & \textbf{Bias} [mg] & \textbf{Noise} [\unit{\micro g/\sqrt{Hz}}] \\ \hline
7200 & 0.004 & 40 & 80 \\ \hline
\end{tabular}
}
\end{table}
Trajectories {1-3,5,6,9-12} were used for training while Trajectories {4,7,8} for testing.
\subsubsection{Legged Robot Dataset}
\noindent 
This publicly available dataset~\cite{frey2026grandtour} was collected using a legged robot equipped ANYmal IMU and CPT7 dual-antenna RTK-GNSS sensors, The dataset comprises a total of 116 minutes of synchronized motion recordings across 49 different trajectories each of trajectories span between 3 minutes to 7 minutes. In this paper, we employed 17 different trajectories as described in Table~\ref{traj_spec_LR}.
\begin{table}[h]
\centering
\caption{Length and time of Legged Robot dataset trajectories}
\label{traj_spec_LR}
\begin{tabular}{|c|c|c|}
\hline
\textbf{Trajectory} & \textbf{Length {[}m{]}} & \textbf{Time {[}sec{]}} \\ \hline
1                   & 198.0                  & 348                     \\ \hline
2                   & 204.0                  & 478                     \\ \hline
3                   & 204.0                  & 478                     \\ \hline
4                   & 137.1                  & 388                     \\ \hline
5                   & 173.5                  & 397                     \\ \hline
6                   & 94.1                   & 228                     \\ \hline
7                   & 168.6                  & 275                     \\ \hline
8                   & 139.4                  & 236                     \\ \hline
9                   & 200.5                  & 444                     \\ \hline
10                  & 179.3                   & 362                     \\ \hline
11                  & 258.5                  & 470                     \\ \hline
12                  & 139.4                  & 236                     \\ \hline
13                  & 200.6                  & 444                     \\ \hline
14                  & 313.6                  & 444                     \\ \hline
15                  & 427.9                  & 960                     \\ \hline
16                  & 244.8                  & 412                     \\ \hline
17                  & 169.5                  & 354                     \\ \hline
\textbf{Average} & \textbf{203.1} & \textbf{409}  \\ \hline
\end{tabular}
\end{table}
Trajectories {1-6,8-15} were used for training while Trajectories {7,16,17} for testing. 
\subsection{Performance Metrics}
\noindent
To evaluate our proposed DDR apporaches relative to the inertial navigation equations of motion, we used two performance matrices:
\begin{enumerate}
    \item \textbf{Position root mean square error (PRMSE)}: The PRMSE of the position compares the estimated position of the dog in the navigation frame with the GNSS GT:
    \begin{equation} \label{eq:rmse}
    \text{PRMSE (m)} = \sqrt{\frac{1}{N} \sum_{k=1}^{N} ||\textbf{p}_k-\hat{\textbf{p}}^{}_k||^{2}}
    \end{equation}
    where $\hat{\textbf{p}}_k$ is the estimated position vector at time k and $\textbf{p}_k$ is the GT position vector at time k.

    \item \textbf{Absolute Distance Error (ADE)}: The ADE of the trajectory compares the estimated and GT trajectory length of the dog:
    \begin{equation} \label{eq:ADE}
    \text{ADE (m)} = |l_{GT}-l_{est}|
    \end{equation}
    where $l_{est}$ is the estimated length trajectory and $l_{GT}$ is the GT length trajectory.
\end{enumerate}
\subsection{Comparative Methods}
\noindent 
To evaluate the performance of our proposed approaches, we compare them against a standard INS baseline. All methods rely exclusively on inertial sensors for the positioning task. The evaluated approaches are:
\begin{enumerate}
    \item \textbf{INS}: A strapdown inertial navigation solution using the body-mounted IMU on the dogs. This baseline solution is obtained using \eqref{eq:INSEOM1}-\eqref{eq:INSEOM3}.
    \item \textbf{MB-DDR}: Our proposed model-based DDR, with the Weinberg step-length estimation,  as described in Algorithm~\ref{alg:ddr}.
    \item \textbf{DL1-DDR}: Our deep learning-aided DDR based on two ResNet backbone network for estimating the dog's velocity and  direction, as presented in Algorithm~\ref{alg:dognet_position_update}.
    \item \textbf{DL2-DDR}:  Our deep learning-aided DDR based on a ResNet backbone network for estimating the dog's velocity and on a transformer encoder network to regress the dog's walking  direction, as shown in Algorithm~\ref{alg:LRE_position_update}.
\end{enumerate}
\subsection{Results: Dog Dataset}
\begin{figure*}[h]
    \centering
    \includegraphics[width=1\linewidth]{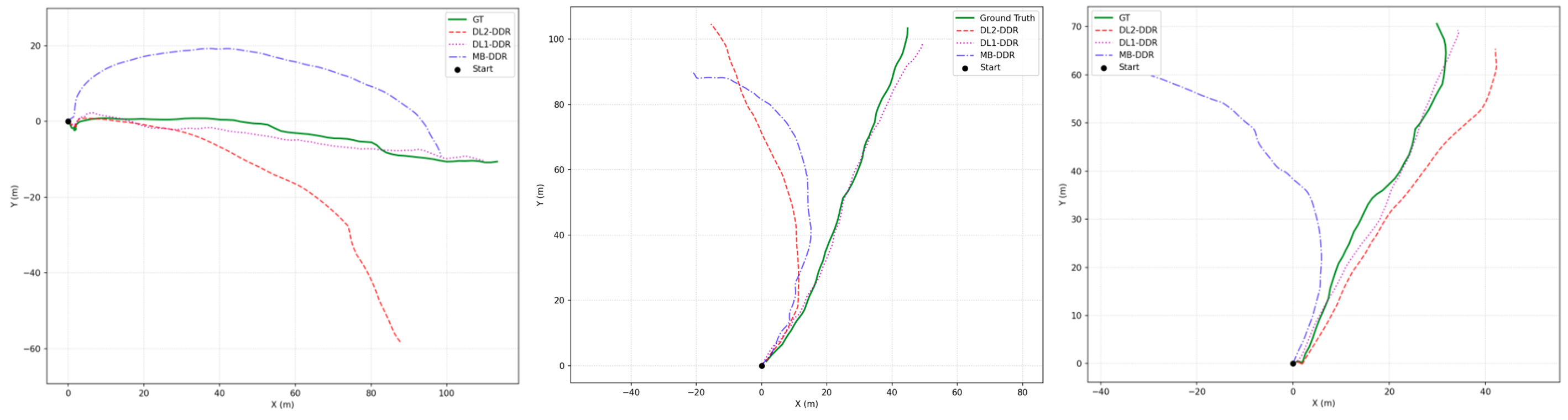}
    \caption{GT trajectory and estimated DDR trajectories for the dog test dataset (Trajectories 4,7, and 8).}
    \label{fig:dog_result}
\end{figure*}

\noindent
At the beginning of each trajectory, we perform a stationary calibration (zero-order) using the first 250 inertial samples ($\sim$2 s) to estimate constant sensor biases and the initial orientation. This assumes that the dog remains stationary during this interval. As expected, the model-based \textbf{INS} approach yielded significant errors across all trajectories, with an average PRMSE of 2400 meters. For the \textbf{MB-DDR}, we used Trajectory {2} to calibrate the Weinberg gain and evaluated it on the remaining trajectories. Table~\ref{tab:res_data1} presents the performance of our proposed approaches for each test trajectory. We note that the\textbf{MB-DDR} achieved consistent performance across the trajectories not included in the table.
\begin{table*}[h]
\centering
\caption{Performance comparison across different DDR trajectories for the dog dataset showing the PRMSE and ADE metrics.}
%\resizebox{\columnwidth}{!}
{
\begin{tabular}{|c|cc|cc|cc|}
\hline
                    & \multicolumn{2}{c|}{\textbf{MB-DDR {[}m{]} (\%)}}                              & \multicolumn{2}{c|}{\textbf{DL1-DDR {[}m{]} (\%)}}                           & \multicolumn{2}{c|}{\textbf{DL2-DDR {[}m{]} (\%)}}                             \\ \hline
\textbf{Trajectory} & \multicolumn{1}{c|}{\textbf{PRMSE}}               & \textbf{ADE}               & \multicolumn{1}{c|}{\textbf{PRMSE}}             & \textbf{ADE}               & \multicolumn{1}{c|}{\textbf{PRMSE}}               & \textbf{ADE}               \\ \hline
\textbf{4}          & \multicolumn{1}{c|}{{28.38  (25.3\%)} }          & {5.08} (4.3\%)          & \multicolumn{1}{c|}{{4.28} (3.9\%)}          & {9.08} (7.6\%)          & \multicolumn{1}{c|}{{30.54} (27.8\%)}          & {9.1} (7.6\%)          \\ \hline
\textbf{7}          & \multicolumn{1}{c|}{{16.18} (13.7\%)}          & {10.46} (13.1\%)        & \multicolumn{1}{c|}{{2.77} (2.3\%)}          & {2.05} (2.6\%)          & \multicolumn{1}{c|}{{21.14} (17.7\%)}          & {2.05} (2.6\%)          \\ \hline
\textbf{8}          & \multicolumn{1}{c|}{{30.52} (38.6\%)}          & {2.13} (3.0\%)          & \multicolumn{1}{c|}{{1.67} (2.4\%)}          & {1.25} (1.8\%)          & \multicolumn{1}{c|}{{6.52} (8.1\%)}            & {1.25} (1.8\%)          \\ \hline
\textbf{Mean}       & \multicolumn{1}{c|}{\textbf{{25.02} (25.9\%)}} & \textbf{{5.9} (6.8\%)} & \multicolumn{1}{c|}{\textbf{{2.91} (2.9\%)}} & \textbf{{4.12} (4.0\%)} & \multicolumn{1}{c|}{\textbf{{19.40} (17.9\%)}} & \textbf{{4.12} (4.0\%)} \\ \hline
\end{tabular}
}
\label{tab:res_data1}
\end{table*}
The experimental results demonstrate that the DL1-DDR method consistently outperforms both the model-based approach and the alternative deep learning configuration across all evaluated metrics. Specifically, DL1-DDR achieved the lowest mean PRMSE of 2.91m (2.9\%), representing a significant improvement over the 25.02 m (25.87\%) recorded for MB-DDR and 19.40m (17.9\%) for DL2-DDR. While MB-PDR shows high sensitivity to trajectory variations peaking at a PRMSE of 30.52m. \\
Interestingly, the ADE metrics reveal a shared performance characteristic between DL1-DDR and DL2-DDR in Trajectories 7 and 8, both yielding identical errors (e.g., 1.25 m in Trajectory 8). However, DL1-DDR’s superior PRMSE suggests it is far more robust against significant positional drifts than DL2-DDR. Overall, the transition from traditional model-based DDR to the DL1-DDR architecture results in a nearly ten-fold reduction in mean positional error, validating the efficacy of deep learning in mitigating the cumulative drift inherent in inertial navigation.\\
\noindent
Figure~\ref{fig:dog_result} presents GT trajectory and estimated DDR trajectories for the dog test dataset (Trajectories 4,7, and 8).
\subsection{Results: Robotic Legged Dog Dataset}
\begin{figure*}[h]
    \centering
    \includegraphics[width=1\linewidth]{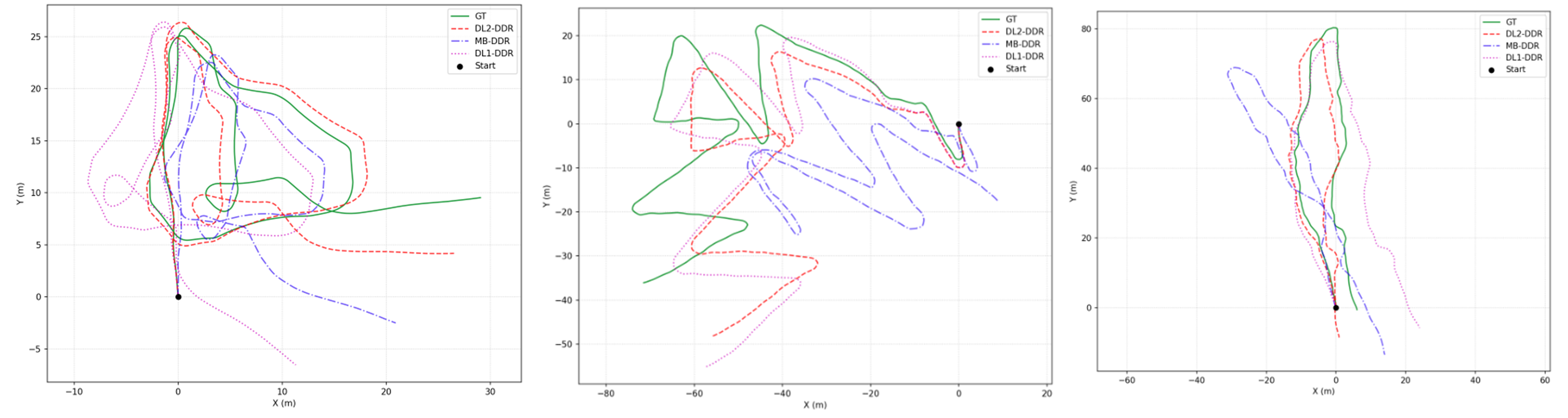}
    \caption{GT trajectory and estimated DDR trajectories for the dog test dataset (Trajectories 7,16, and 17).}
    \label{fig:robot_result}
\end{figure*}

\noindent
In this dataset no prior prepossessing was required. As in the dog dataset, the model-based \textbf{INS} approach yielded significant errors across all trajectories, with an average PRMSE over 10km. For the \textbf{MB-DDR}, we used Trajectory {12}  to calibrate the Weinberg gain and evaluated it on the remaining trajectories. Table~\ref{tab:res_data2} presents the performance of our proposed approaches for each test trajectory. We note that the\textbf{MB-DDR} achieved consistent performance across the trajectories not included in the table.
\begin{table*}[h]
\centering
\caption{Performance comparison across different DDR trajectories for the legged dog dataset showing the PRMSE and ADE metrics.}
%\resizebox{\columnwidth}{!}
{
\begin{tabular}{|c|cc|cc|cc|}
\hline
\multirow{2}{*}{\textbf{Trajectory}} & \multicolumn{2}{c|}{\textbf{MB-DDR {[}m{]} (\%)}}                              & \multicolumn{2}{c|}{\textbf{DL1-DDR {[}m{]} (\%)}}                             & \multicolumn{2}{c|}{\textbf{DL2-DDR {[}m{]} (\%)}}                            \\ \cline{2-7} 
                                     & \multicolumn{1}{c|}{\textbf{PRMSE}}              & \textbf{ADE}                & \multicolumn{1}{c|}{\textbf{PRMSE}}              & \textbf{ADE}                & \multicolumn{1}{c|}{\textbf{PRMSE}}             & \textbf{ADE}                \\ \hline
\textbf{7}                           & \multicolumn{1}{c|}{{6.47} (4.6\%)}           & {20.08} (14.9\%)         & \multicolumn{1}{c|}{{8.71} (4.3\%)}           & {4.06} (2.0\%)           & \multicolumn{1}{c|}{{2.11} (1.5\%)}          & {4.06} (2.0\%)           \\ \hline
\textbf{16}                          & \multicolumn{1}{c|}{{57.4} (23.5\%)}          & {20.29} (8.3\%)          & \multicolumn{1}{c|}{{10.81} (4.4\%)}          & {17.98} (7.4\%)          & \multicolumn{1}{c|}{{8.70} (3.6\%)}          & {17.98} (7.4)          \\ \hline
\textbf{17}                          & \multicolumn{1}{c|}{{13.92} (8.2\%)}          & {9.51} (5.6\%)           & \multicolumn{1}{c|}{{19.24} (11.3\%)}         & {7.33} (4.3
\%)           & \multicolumn{1}{c|}{{4.94} (2.9\%)}          & {7.33} (4.3\%)           \\ \hline
Mean                                 & \multicolumn{1}{c|}{\textbf{{25.93} (12.1\%)}} & \textbf{{16.62} (9.6\%)} & \multicolumn{1}{c|}{\textbf{{12.92} (6.7\%)}} & \textbf{{9.79} (13.7\%)} & \multicolumn{1}{c|}{\textbf{{5.25} (2.7\%)}} & \textbf{{9.79} (13.7\%)} \\ \hline
\end{tabular}
}
\label{tab:res_data2}
\end{table*}
Based on the results presented in Table~\ref{tab:res_data2}, the performance evaluation across Trajectories 7, 16, and 17 highlights a significant shift in accuracy compared to previous tests, with DL2-DDR emerging as the most precise model. While MB-DDR continues to struggle with high error rates, reaching a peak PRMSE of 57.4m in Trajectory 16, the deep learning models demonstrate superior adaptation to the diverse movement patterns. The DL2-DDR architecture achieved the lowest mean PRMSE of 5.25m (2.7\%), effectively halving the error of the DL1-DDR 12.92m. Notably, while DL1-DDR and DL2-DDR share identical ADE values across all individual trajectories, the PRMSE metric reveals that DL2-DDR is significantly more capable of minimizing peak positional deviations. This suggests that while both deep learning configurations may follow a similar performance, DL2-DDR provides a more stable and geometrically accurate reconstruction of the legged dog's path, particularly in complex scenarios where the model-based approach's mean error exceeds 25m.
\noindent
Figure~\ref{fig:robot_result} presents GT trajectory and estimated DDR trajectories for the legged dog test dataset (Trajectories 7,16, and 17).
\subsection{Summary}
\noindent
Regardless of whether the subject is a biological dog or a robotic legged dog, DL-DDR models significantly outperform the MB-DDR approach and of course the INS solution.  MB-DDR consistently exhibits the highest PRMSE errors, indicating that kinematic models struggle with the cumulative drift inherent in long-term dog tracking.\\
\noindent
Across both tables, DL1-DDR and DL2-DDR often share the exact same ADE for specific trajectories (e.g., Trajectory 7 and 8 in Table 1; Trajectories 7, 16, and 17 in Table 2). This suggests that while their internal error distributions differ, their average estimated trajectory remains  similar. \\
\noindent
To further validate the robustness of the proposed DL-DDR architectures, an ablation study was conducted by randomly reassigning the trajectories between the training and testing datasets. Despite these stochastic variations in the data distribution, the performance metrics remained remarkably consistent across multiple iterations. Both DL1-DDR and DL2-DDR maintained their  accuracy with mean PRMSE and ADE values fluctuating by negligible margins. This stability indicates that the models have successfully learned generalized motion features—such as gait patterns and inertial signal variation rather than merely overfitting to specific trajectories. The results confirm that the performance gains observed in Tables 1 and 2 are a result of the architectural design of the DL models and are not biased by the specific partitioning of the data.

\section{Conclusions}\label{sec:conc}
\noindent
In this paper  we present a framework for dog dead reckoning suitable for both biological dogs and robotic legged dogs. To this end,  we proposed three methods including a model-based DDR and two neural-aided DDR. We validated our methods using two distinct datasets: 1) Our dataset recorded with two dog subjects and 2) a publicly available dataset that was collected using a legged robot.\\
\noindent 
The experimental results highlight the limitations of classical inertial integration for this application. 
While the model-based DDR method substantially improves performance by introducing step-based updates, the heading angle was not estimated accurately, leading to a degraded reconstructed trajectory. In contrast, our neural-aided approaches, DL1-DDR and DL2-DDR produce more robust short-term velocity and direction estimates, which significantly reduce trajectory drift across both datasets. This improvement over the MB-DDR is attributed to their  inherent ability to denoise sensor signals, model complex relationships, and cope with uncertainty in the inertial data. \\
\noindent Future work will focus on expanding the dataset to include additional dogs, environments, and motion patterns in order to improve robustness and generalization. Furthermore, we plan to incorporate magnetometer measurements for further reducing drift and improving overall localization performance.\\
In conclusion, our DDR approaches offer a lightweight and low-cost solution for both biological dogs and legged robotic dogs. These methods overcome the high drift characterizing of pure inertial navigation,  enabling the achievement of mission-specific objectives. Furthermore, by providing accurate positioning for legged robots, our methods facilitate autonomy on low-cost hardware and contribute to effective collision avoidance. To ensure reproducibility, provide a standardized benchmark for the community, and stimulate further research, the codebase implementing the described frameworks and our new dataset have been  made publicly available at a \href{https://github.com/ansfl/DDR}{GitHub repository}.

\bibliographystyle{ieeetr}
\bibliography{bio}

\end{document}